\documentclass[lettersize,journal]{IEEEtran}
\usepackage{soul, color, xcolor}
\usepackage{amsthm,amsmath,amssymb}
\usepackage{mathrsfs}
\usepackage{dutchcal}
\usepackage{amsfonts}
\usepackage{algorithmic}
\usepackage{algorithm}
\usepackage{array}
\usepackage[caption=false,font=normalsize,labelfont=sf,textfont=sf]{subfig}
\usepackage{textcomp}
\usepackage{stfloats}
\usepackage{url}
\usepackage{verbatim}
\usepackage{graphicx}
\usepackage{tabularx}
\usepackage{wrapfig}
\usepackage{cite}
\usepackage{hyperref}
\usepackage{float}
\usepackage[marginal]{footmisc} 

\begin{document}

\title{Revolutionizing Turn-by-Turn Navigation with Cloud-Edge Deep Learning}

\author{%
  Yiming Yang$^*$, Hao Fu$^*$, Fanxiang Zeng$^*$, Xikai Yang, Yue Liu, and Ning Guo \\
  \thanks{$^*$ These authors contributed equally to this work and should be considered co-first authors.} 
  AMap, Alibaba Group, Beijing, China \\
  Email: \{sachiel.yym, fh265565, fanxiang.zfx, xikai.yxk, yue.liu, ning.guo\}@alibaba-inc.com%
}




\maketitle

\begin{abstract}
Turn-by-turn (TBT) navigation systems are integral to modern driving experiences, providing real-time audio instructions to guide drivers safely to destinations. However, existing audio instruction policy often relies on rule-based approaches that struggle to balance informational content with cognitive load, potentially leading to driver confusion or missed turns in complex environments. To overcome these difficulties, we first model the generation of navigation instructions as a multi-task learning problem by decomposing the audio content into combinations of modular elements. Then, we propose a novel deep learning framework that leverages the powerful spatiotemporal information processing capabilities of Transformers and the strong multi-task learning abilities of Mixture of Experts (MoE) to generate real-time, context-aware audio instructions for TBT driving navigation. A cloud-edge collaborative architecture is implemented to handle the computational demands of the model, ensuring scalability and real-time performance for practical applications. Experimental results in the real world demonstrate that the proposed method significantly reduces the yaw rate (the proportion of vehicles deviating from navigation routes) compared to traditional methods, delivering clearer and more effective audio instructions. This is the first large-scale application of deep learning in driving audio navigation, marking a substantial advancement in intelligent transportation and driving assistance technologies.
\end{abstract}

\begin{IEEEkeywords}
Deep Learning, Turn-by-Turn Navigation, Cloud-Edge Collaboration.
\end{IEEEkeywords}

\section{Introduction}
Navigation systems in the era of mobile internet have improved the driving experience by providing drivers with route information and directions in real-time via visual and audio instruction on the navigation terminal \cite{yang2024effects}. Compared with visual information, drivers tend to rely more on audio instructions for the sake of driving safety \cite{zhong2022address}. However, current turn-by-turn (TBT) driving navigation \cite{sara2023exploring} often finds it challenging to strike a balance between yaw rate (the proportion of vehicles deviating from navigation routes), play timing, and play density for audio instruction. Overly complex audios can increase the driver's cognitive load, making it difficult to understand information quickly and safely \cite{dalton2013driving}. Conversely,implistic audio may fail to provide sufficient guidance for navigating complex intersections and intricate road networks, leading to yaw and compromised safety \cite{LARGE201469.e1}.

The principal difficulty in generating real-time navigation instructions lies in how to play the accurate and complete audio at the correct time. 
Current methods typically use rule-based policies or pre-defined configuration tables \cite{jensen2010studying,LARGE201469.e1,yang2021experimental} that lack the flexibility, making it easy to fall into the seesaw effect between yaw rate, play timing, and audio density. These approaches may result in navigation instructions that are either too general, failing to convey critical information, or too verbose, overloading the driver with unnecessary details.

To address these challenges, we propose a novel pipeline for navigation instruction generation: firstly, decomposing the audio context into modular elements, then the navigation instruction model in charge of element recall, play timing, and order selection, and finally generating coherent speech via text-to-speech (TTS) module \cite{kaur2023conventional}. This pipeline transforms the complex task of TBT driving navigation into manageable components, allowing for precise and adaptable instructions that effectively balance informational content with cognitive load.

Building upon this formalization, we introduce the first deep learning framework utilizing sequence models for real-time, context-aware audio instruction generation in practical TBT driving navigation. Our method captures the complex dependencies and variations inherent in driving scenarios. Leveraging sequence modeling, our approach effectively handles complex intersections and effectively reduces yaw rate.

Our main contributions are as follows:
\begin{itemize}
    \item \textbf{The First Deep Learning Based TBT Driving Navigation:} To the best of our knowledge, we are the first to implement the deep learning based TBT navigation policy for practical applications, utilizing sequence models to capture spatiotemporal dependencies and address the seesaw effect between yaw rate, play timing, and audio density.
    \item \textbf{Cloud-Edge Collaborative Architecture:} We implement a cloud-edge collaborative architecture to handle the real-time computational demands of navigation instruction model, ensuring scalability and real-time performance for large-scale service deployment.
    \item \textbf{Data-Driven Paradigm for TBT Optimization:} We introduce the data-driven paradigm for optimizing TBT driving navigation, which shifts from rule-based to data-driven optimization results in continuous performance improvements.
\end{itemize}

\section{Preliminaries and Related Work}\label{sec:preliminaries}
\subsection{Preliminaries}
TBT driving navigation refers to a navigational aid system that provides step-by-step instructions to drivers, guiding them from a starting location to the destination. This system utilizes real-time driving data, often incorporating Global Positioning System (GPS) technology, to help with wayfinding problems during driving \cite{schwering2017wayfinding,sara2023exploring}. Instructions are typically delivered via audio prompts and visual cues, indicating when and where to make turns, lane changes, and other operations.
\begin{figure}[htbp]
\centering
\includegraphics[width=9cm]{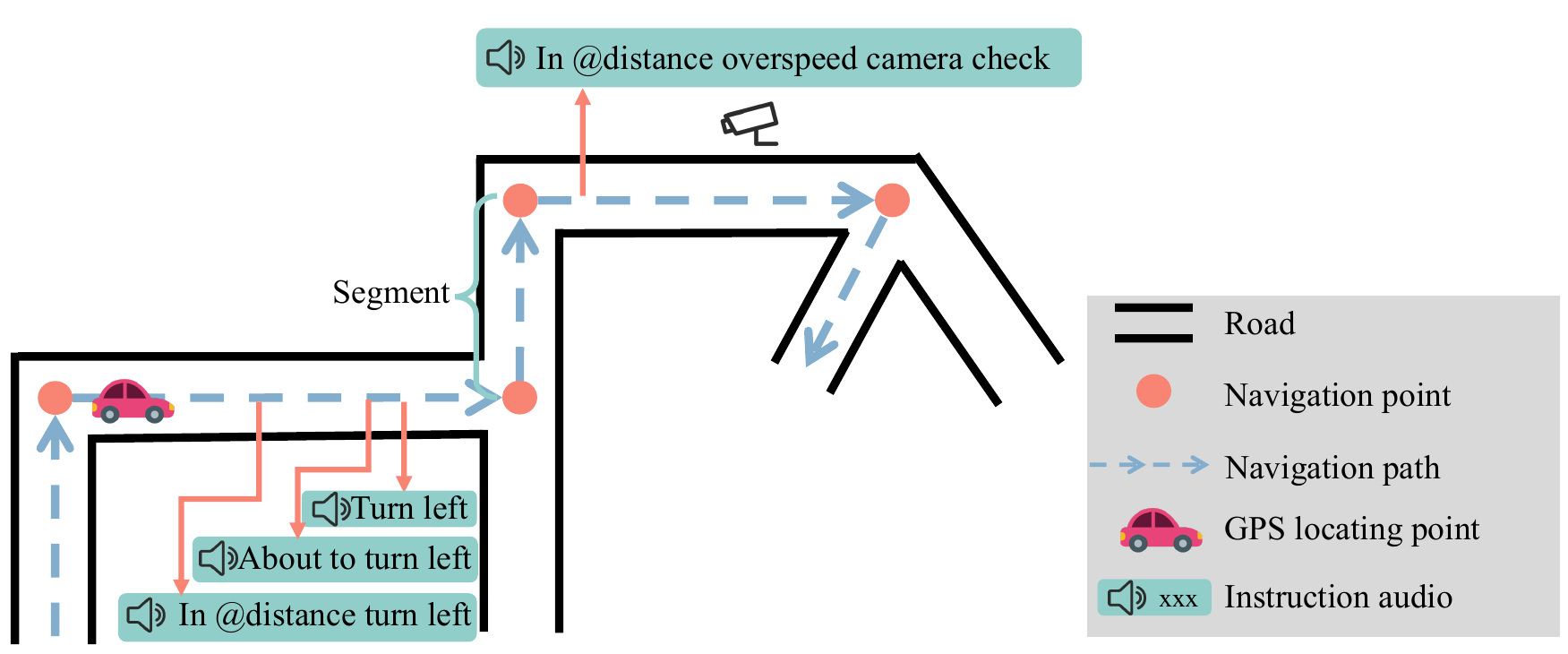}
\caption{\textbf{TBT driving navigation.} The figure displays the essential cue content and key terminology used in the TBT driving navigation, along with the paradigm of the audio instruction while the car moves through the path.}
\label{fig:tbt_env}
\end{figure}

The basic flow of audio instructions is presented in Figure \ref{fig:tbt_env}, illustrating key concepts within the TBT driving navigation we have defined. The navigation point represents the location where the navigation system expects the driver to make a steering to avoid yawing, typically at a fork in the road. The directed connecting line between two neighboring navigation points is called as segment. All segments form the navigation path. The primary goal of the audio instruction policy during navigation is to select the appropriate content at the proper time to prevent the driver from yawing or violating traffic rules, where yawing refers to the driver deviating from the planned route provided by the navigation system, which usually means that the instruction content is wrong or poorly timed, causing the driver to go the wrong way.

The audio content must follow a standardized approach and be brief to ensure quick comprehension by the driver. Consequently, unlike other language generation tasks, when generating instruction audio, sentence diversity is not a consideration. This allows us to organize instructional audio content using key elements and consistent connectors, where the key information entities related to driving are called elements. We define a set of elements $\mathcal{E}_\text{elem}=\{\mathcal{E}_{\text{action}},\mathcal{E}_{\text{info}}\}$ that encapsulate the key information to be conveyed through audio instructions. These elements are categorized into:
\begin{itemize}
\item \textbf{Action Elements} $\mathcal{E}_{\text{action}}$: Elements that require the driver needs to turn the wheel following the audio instruction, such as ``turn left" or ``merge right."
\item \textbf{Info Elements} $\mathcal{E}_{\text{info}}$: Elements that provide information without requiring to turn the wheel, such as ``speed camera ahead" or ``exceeding speed limit, reduce your speed".
\end{itemize}
Under this paradigm, each audio instruction contains audio content, play timing, and play order in the segment. Based on this, we structure an instruction audio as: the elements that need to be revealed, the order within the segment, and the play timing of the audio.
Considering the cognitive load of the driver, each audio content may contain multiple
action elements, but at most one info element.
The play timing is indicated by the relative distance from the audio playing position to the navigation point. The term ``order" refers to the position of the current audio in the segment after all audios are sorted by play position. It is related to the selection of the connectors. Figure \ref{fig:tbt_env} shows the connections of play order in the segment and the organization of audio content through an example: 3 green boxes represent 3 different instruction audios, each with the same element ``turn left". However, the connectors for the element vary depending on the play order, resulting in corresponding content changes.

\subsection{Related Work}

\textbf{Traditional Navigation Methods.}
Traditional TBT navigation systems typically follow a pipeline of positioning, map matching, path planning, and instruction triggering. Hidden Markov Models (HMMs) have long served as the standard algorithm for map matching, which maps noisy GPS sequences to the most probable road segments \cite{newson2009hidden, mor2021systematic}. HMMs treat driving states as hidden variables and observed GPS coordinates as emissions, enabling probabilistic inference under topological constraints. However, when it comes to instruction triggering---determining ``when to speak'' and ``what to say''---traditional systems predominantly rely on geometric rules. For instance, voice instructions are triggered when the vehicle enters a predefined radius (trigger radius) near an intersection and the map matching state confirms an upcoming turn \cite{yang2021experimental, bian2021influence}. While HMMs effectively solve the ``where is the car'' problem, they remain limited to the Markovian assumption and cannot capture long-range temporal dependencies. Rule-based triggering methods, though interpretable, lack the flexibility to determine optimal play timing under dynamic traffic conditions, often resulting in the ``seesaw effect'' between yaw rate, audio density, and play timing.

\textbf{Deep Learning in Navigation.}
Significant advances have been made in applying deep learning to navigation-related tasks. For positioning and perception, researchers have designed novel objectives to improve positioning accuracy in GPS-denied environments \cite{liu2021design}, and developed quality control measures for multisensor systems to enhance vehicle navigation in urban areas \cite{wang2023measurement}. In the realm of visual navigation, methods like Navigation World Models \cite{bar2024navigationworldmodels, shah2023gnm} and semantic segmentation-driven navigation \cite{krishna2024machine} have demonstrated robust capabilities in wayfinding and exploration. End-to-end driving approaches \cite{codevilla2018end} use images and high-level navigation commands to directly predict vehicle control signals, while Vision-and-Language Navigation (VLN) research \cite{anderson2018vision, gu2022vision} has explored instruction-following in simulated environments. Deep reinforcement learning has been applied to autonomous vehicle path planning \cite{kiran2021deep}, and sequence models like LSTMs and Transformers have been employed for trajectory prediction \cite{zhao2019multi, giuliari2021transformer}. However, these efforts primarily focus on robotic navigation, autonomous driving perception, or path planning---none directly addresses TBT audio policy for human drivers.

Turn-by-Turn (TBT) audio navigation for individuals with low vision has been extensively studied. For example, Navigate-Me \cite{dissanayake2021navigate} utilizes secure voice authentication for indoor navigation, while systems such as iSAFE NAVigation Vest \cite{miranda2023enhancing} integrate audio with collision detection to enhance safety. Other works leverage object detection and depth sensing \cite{malkan2021navigation} or augmented reality audio cues \cite{chi2022enabling} to provide intuitive guidance for visually impaired users. These approaches highlight the potential of well-crafted navigation policies to improve situational awareness and decision-making. However, these studies predominantly target accessibility needs and indoor or pedestrian navigation scenarios, offering limited insights into the challenges of generating real-time, context-aware navigation instructions for human driving.

Despite its critical role in driving safety, research into audio navigation policies for human driving has received less attention. While studies like Wunderlich et al. \cite{wunderlich2023landmark} have explored landmark-augmented audio navigation to enhance spatial awareness, and other works \cite{yang2021experimental, bian2021influence} have investigated the effects of prompt timing and message content, these efforts remain grounded in static, rule-based design principles. Consequently, they struggle to address the ``seesaw effect'' between yaw rate, audio density, and play timing, where attempts to optimize one metric often adversely impact the others.

In contrast, our work is the first to apply deep learning specifically to TBT audio policy---the scheduling strategy for generating and timing voice instructions for human drivers. By leveraging sequence modeling and multi-task learning, our approach can capture complex spatiotemporal dependencies and generate flexible, context-aware navigation instructions, transcending the limitations of traditional rule-based methods.

\begin{figure*}[htbp]
\centering
\includegraphics[width=18cm]{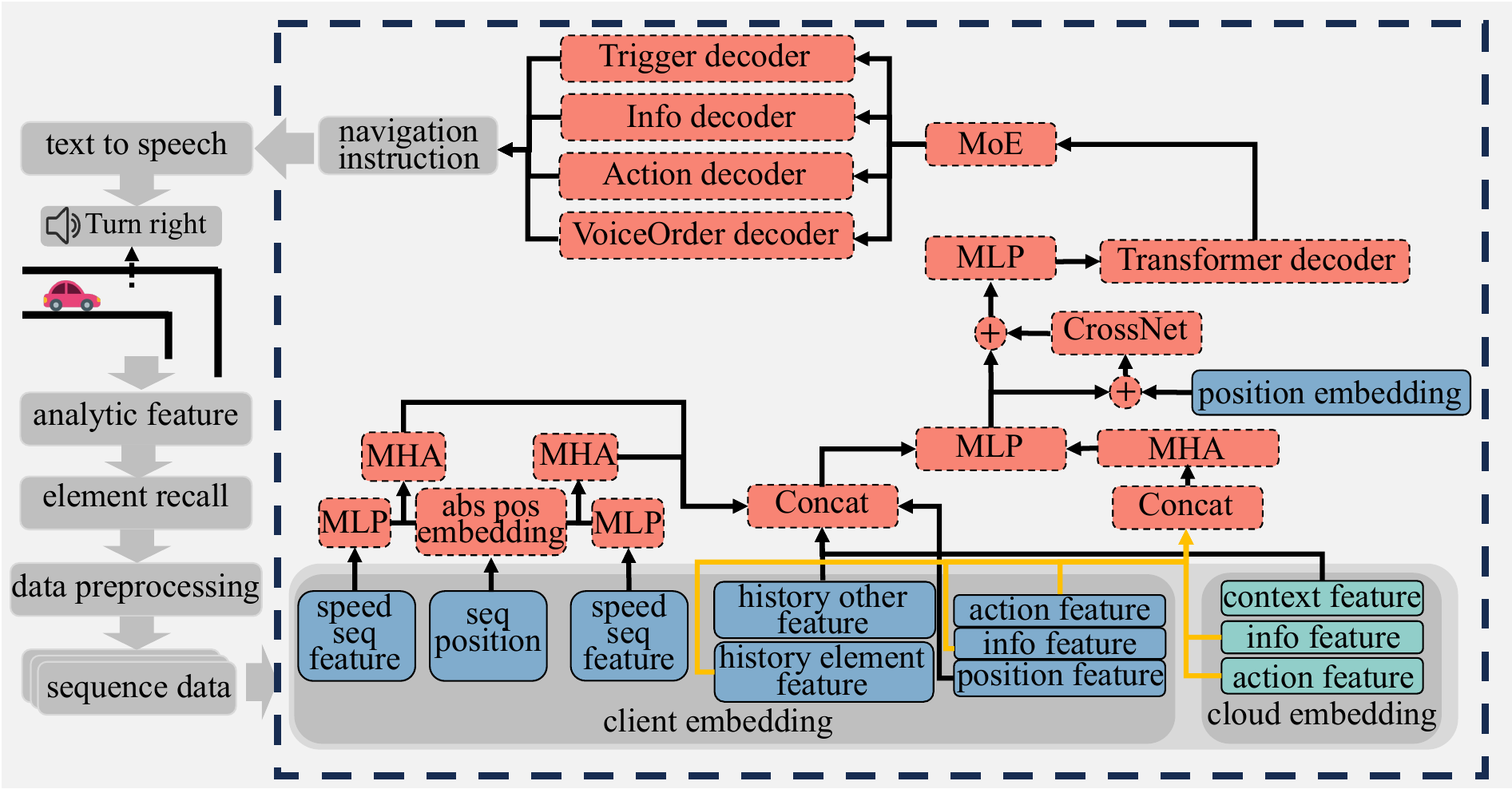}
\caption{\textbf{Overview of TBT navigation instruction model.} In the dashed box is the proposed navigation instruction model. The dashed red box represents the model component, the solid blue box represents the edge-side data embedding, and the solid green box represents the cloud-side data embedding. To the left of the dashed box is the engineering design for model real-time playing and data preparation. Abbreviations: MoE means Mixture of Experts, MLP means Multi-Layer Perceptron, MHA means Multi-Head Attention.}
\label{fig:overview}
\vspace{-0.56cm}
\end{figure*}

\section{Method}
In this section, we delve into the details of our approach, covering the following aspects: Firstly, Section \ref{subsec:formalization} formalizes the audio instruction problem as multi-task learning. Then, Section \ref{subsec:overview} introduces the sequence model for audio instruction, along with the cloud-edge collaboration for model deployment. Finally, Section \ref{subsec:training} outlines the model training.

\subsection{Problem Formalization} \label{subsec:formalization}

To address the challenges in generating real-time, context-aware instructions, we model the TBT navigation instruction as a multi-task learning (MTL) problem. The inherent complexity of this task, which requires concurrently determining the appropriate content, its linguistic structure, and precise delivery timing, renders an MTL framework a particularly suitable paradigm. The principle of leveraging MTL to manage multifaceted objectives and enhance system performance is recognized in related intelligent transportation domains. For instance, MTL has been employed to bolster vehicular navigation system performance during GPS outages \cite{9212618}, to model sophisticated driving strategies for autonomous vehicles at intersections \cite{liu2023multi}, and to improve an autonomous system's ability to react to traffic signals \cite{ishihara2021multi}. In a similar spirit, adopting an MTL approach for TBT audio instructions enables our model to concurrently optimize the multiple, interdependent components essential for generating effective guidance.

As presented in Section \ref{sec:preliminaries}, the instructions within each segment have a strong spatiotemporal correlation, so segments are selected as the granularity for modeling driving scenarios oriented towards instruction.
We sample features for the navigation instruction model in segments at 1-second intervals: $x_{t_1}, x_{t_2}, \dots, x_{t_T}$, where $T$ is the total number of time steps the car passes the segment.

Considering the importance of feature sequences for the navigation instruction task, we aim to learn a function that maps the input sequence data to several outputs to compose instructions:

\begin{equation}
f: \mathcal{X}_{t} \rightarrow \{ y_{\text{trigger}},\ y_{\text{action}},\ y_{\text{info}},\ y_{\text{vo}} \},
\end{equation}

where $\mathcal{X}_{t}=\{\mathcal{x}_{t-n},...,\mathcal{x}_{t-1},\mathcal{x}_t\}$ represents the input sequence features, $n$ is the sequence length. 
Let $d_{\text{np}}$ denote the distance from the current driver's position to the navigation point, and $d_{\text{pp}}$ denote the distance from the expected audio play position to the navigation point.
$y_{\text{trigger}}$ represents the ratio $y_{\text{trigger}}=\frac{d_{\text{pp}}}{d_{\text{np}}}$, indicating the audio play timing.
$y_{\text{action}} \in \{0,1\}^{|\mathcal{E}_{\text{action}}|}$ is a binary vector indicating which action elements should be included in the audio. $y_{\text{info}} \in \{0,1\}^{|\mathcal{E}_{\text{info}}|}$ is a binary vector indicating which info elements should be included. $y_{\text{vo}} \in \{1,2,\dots,O\}$ represents the play order of the audio instruction within the segment, where $O$ is the maximum number of possible orders.

By formalizing the problem in this way, we can model the mapping function from input features to outputs via deep neural networks based on maximum likelihood estimation to fit the distribution of high-quality data to learn a better audio instruction policy in TBT navigation.

\subsection{Sequence modeling}\label{subsec:overview}

Figure \ref{fig:overview} shows our sequence model and its application framework in the TBT navigation system. The model adopts a cloud-edge cooperative architecture, considering scalability, real-time computation, and resource optimization. The responsibility of the cloud side is to embed features that are relatively static in the segment. The edge side is responsible for embedding features that have high real-time requirements and then performing model inference on the edge side along with the feature embedding sent down from the cloud. A detailed description and analysis of the advantages of the cloud-edge architecture are provided in Section \ref{appendexsec:cloud-edge}.

Since the current instruction content generation is strongly correlated with the historical play within the segment, we combine the current timestep features with previous $n-1$ moments in chronological order to form the input sequence features $\mathcal{X}_{t}=\{\mathcal{x}_{t-n},...,\mathcal{x}_{t-1},\mathcal{x}_t\}$. If there are less than $n$ instructions in the segment, zero padding is applied to the missing portion of the sequence. Unlike sequence labeling tasks, the navigation instruction encounters challenges in calculating sequence loss due to the inability to predetermine the input sequence \cite{huang2015bidirectionallstmcrfmodelssequence}. Therefore, we have tailored a sequence model for TBT navigation by integrating the spatiotemporal information processing capabilities of the Transformer.

Upon completion of model inference, the predicted timing by the trigger head is used to assemble a complete sentence by combining the inferred action elements via the action head and information element via the info head with connecting word templates based on the voice order head. The sentence is then converted into speech via a Text-to-Speech (TTS) module and ultimately plays in the driver's navigation terminal. This process is iteratively executed throughout the entire path, constituting a comprehensive TBT navigation system.

It should be noted that the action head and the info head are responsible for recalling elements required in the current play content, while all candidate elements are given to the model as input features. The candidate elements in the input features are generated per segment by the scheduling unit based on the current road graph and path planning information.

The model architecture illustrated within the dashed box in Figure \ref{fig:overview} can be broadly divided into 4 levels: the \textit{Feature Encoder}, the \textit{Deep CrossNet} \cite{zheng2018crossnet}, the \textit{Transformer Decoder} \cite{brown2020languagemodelsfewshotlearners}, and the \textit{Mixture of Experts (MoE) Prediction Layer}:

Initially, the sequence data undergoes \textit{Feature Encoder}. All element-related features are embedded and concatenated, which are indicated by the orange arrows in Figure \ref{fig:overview}. The element embeddings are then fully encoded and mixed through the Multi-Head Attention (MHA). After that, the mixed element embedding is encoded via a Multi-Layer Perceptron (MLP) along with other input features. The \textit{Feature Encoder} transforms the high-dimensional sparse feature representations into low-dimensional dense vectors while capturing and preserving the intrinsic structure and semantic information of the data, facilitating subsequent model processing.

Subsequently, the encoded element features are combined with position embeddings. Different from the existing position embedding method \cite{Vaswani2017,Devlin2019BERTPO,2024RoFormer}, we integrate domain-specific knowledge to transform the conventional absolute position embedding into a combination of temporal sequence encoding and spatial semantic encoding. Temporal sequence encoding targets each effective time slice in the sequence, performing reverse indexing and learning through a position embedding matrix. As for spatial semantic encoding, considering that the density of audio instruction increases as the car approaches the next navigation point $d_{\text{np}}$, a distance-based weight discount factor $\gamma \in (0, 1]$ is applied, which modulates the attention weights in the position embedding. The greater $d_{\text{np}}$, the smaller $\gamma$.

We design the position embedding as in Equation \ref{equ:gamma}, where features with smaller distances to the next navigation point $d_{np}$ will receive more attention during the multi-head attention computation. This is essentially an additional inductive bias that we provide to the multi-head attention computation based on domain knowledge. A similar design has been verified as valid in past research on transformers \cite{9807399}.
\begin{equation}
\gamma = \left\{
\begin{array}{lllll}
\frac{1}{2^{\lfloor \frac{d_{\text{np}}}{50}-1 \rfloor}}, &  0\le d_{navi} \le 300\\
\frac{1}{2^{\lfloor \frac{d_{\text{np}}}{100}+2 \rfloor}}, &  300\leq d_{navi} \le 600\\
\frac{1}{2^{\lfloor \frac{d_{\text{np}}}{200}+6 \rfloor}}, &  600\leq d_{navi} \le 1000\\
\frac{1}{2^{\lfloor \frac{d_{\text{np}}}{500}+9 \rfloor}}, &  1000\leq d_{navi} \le 3000\\
\frac{1}{2^{14}}, &  d_{\text{np}} \geq 3000
\end{array}
\right.\label{equ:gamma}
\end{equation}

The data are then processed through the \textit{Deep CrossNet}, which constructs and learns high-order cross-feature combinations. The data is then merged through a residual connection and further encoded by an MLP before the next part.

The \textit{Transformer Decoder} is designed to exploit the spatiotemporal coupling information inherent in sequential data. Each time slice in the data can establish associations with other time slices in the sequence, rather than relying solely on adjacent time slice data. 
This design overcomes the limitations of recurrent architectures like LSTMs - which process sequences incrementally with constrained memory horizons - by explicitly modeling long-range dependencies critical for audio instruction tasks.
By computing the attention map, the model adaptively captures the rich semantic information within the sequence. 
We choose a Decoder-only architecture for the spatiotemporal data processing module because its self-supervised training paradigm naturally aligns with predicting current audio based on sequence features. This autoregressive framework allows the model to effectively learn temporal dependencies within the sequence data, enhancing its capacity to generate accurate and context-aware audio instructions.

Finally, the sequence data processed by the \textit{Decoder only Transformer Decoder} is fed into the \textit{MoE Prediction Layer} for multi-task learning. This layer simultaneously learns to predict 4 sub-tasks necessary for generating an instruction audio: the audio trigger time, the action-type elements included in the audio, the information-type element included in the audio, and the audio play order within the segment. The underlying shared features are learned using the MoE. Through different combinations of these expert networks, each subsequent sub-task head can efficiently focus on the features most pertinent to its specific requirements.

\subsection{Model Training}\label{subsec:training}
As illustrated in Figure \ref{fig:overview}, the outputs of the TBT audio instruction model are divided into 4 sub-task outputs: trigger, action, info, and voice order. Each sub-task has a unique loss function tailored to its specific prediction task:

The trigger decoder is responsible for predicting the normalized play timing of the audio instruction, with its scalar output $\hat{y}_{\text{trigger}} \in [0,1]$. The mean squared error (MSE) is employed as the loss function:
\begin{equation}
\mathcal{L}_{\text{trigger}}= (y_{\text{trigger}} - \hat{y}_{\text{trigger}})^2,
\end{equation}
where $y_{\text{trigger}}$ is the audio play timing label.

The action decoder is responsible for predicting the action-type elements that should be included in the audio instruction from the available action elements in the segment. Since a single audio instruction may contain multiple action-type elements, the action decoder outputs a prediction vector with a length equal to the number of action-type elements. Each probability prediction value $p_i \in [0,1]$ in the prediction vector corresponds to an action element $i$. If $p_i > 0.5$, the action element is included in the audio; otherwise, it is excluded. The loss function is defined as follows:
\begin{equation}
\mathcal{L}_{\text{action}}= (\boldsymbol{\mathcal{y}}_{\text{action}} - \hat{\boldsymbol{\mathcal{y}}}_{\text{action}})^2,
\end{equation}
where $\boldsymbol{\mathcal{y}}_{\text{action}}$ is the label vector for action elements. The position corresponding to the element contained in the audio is 1, otherwise 0.

The info decoder is tasked with predicting the information-type elements that should be included in the audio instruction from the available information elements in the segment. Since one audio instruction can contain at most one information-type element, cross-entropy loss is employed:
\begin{equation}
\mathcal{L}_{\text{info}}= -\sum_{i=1}^{25}\boldsymbol{\mathcal{y}}_{\text{info},i} \log \hat{\boldsymbol{\mathcal{y}}}_{\text{info},i},
\end{equation}
where 25 is the total number of information-type elements plus one, with the additional position indicating the probability that no information-type element is included in the audio instruction. $\boldsymbol{\mathcal{y}}_{\text{info}}$ is the one-hot label representing the information-type element included in the audio instruction. Here, $\boldsymbol{\mathcal{y}}_{\text{info},i}$ denotes the $i$-th element of the ground truth one-hot label vector, and $\hat{\boldsymbol{\mathcal{y}}}_{\text{info},i}$ denotes the corresponding predicted probability for the $i$-th class.

The voice order decoder is responsible for predicting the play order of the audio instruction within the segment. It uses one-hot encoding to classify the order into five categories, ranging from 0 to 4, where 0 indicates that the current audio instruction should not be played, and 1 to 4 represents the play order of the audio relative to the endpoint of the navigation segment. Cross-entropy loss is employed:
\begin{equation}
\mathcal{L}_{\text{vo}}= -\sum_{i=0}^{4}\boldsymbol{\mathcal{y}}_{\text{vo},i} \log \hat{\boldsymbol{\mathcal{y}}}_{\text{vo},i},
\end{equation}
where $\boldsymbol{\mathcal{y}}_{\text{vo}}$ is a one-hot encoded label indicating the instructions' play order within the segment. $\hat{\boldsymbol{\mathcal{y}}}_{\text{vo}}$ represents the predicted play order of the current instruction within the segment. Similarly, $\boldsymbol{\mathcal{y}}_{\text{vo},i}$ denotes the $i$-th element of the ground truth one-hot label for voice order, and $\hat{\boldsymbol{\mathcal{y}}}_{\text{vo},i}$ denotes the corresponding predicted probability.

To address the issue of varying learning difficulties across different sub-tasks in multi-task training and to avoid subpar performance in certain sub-tasks, the total loss function is computed using the geometric mean:
\begin{equation}
\mathcal{L}_{\text{total}} = \left( \mathcal{L}_{\text{trigger}} \cdot \mathcal{L}_{\text{action}} \cdot \mathcal{L}_{\text{info}} \cdot \mathcal{L}_{\text{vo}} \right)^{\frac{1}{4}},
\end{equation}
where $\mathcal{L}_{\text{trigger}}$, $\mathcal{L}_{\text{action}}$, $\mathcal{L}_{\text{info}}$, and $\mathcal{L}_{\text{vo}}$ are the individual loss functions for the trigger, action, info, and voice order decoders, respectively. The geometric mean ensures a balanced contribution from each sub-task, mitigating the risk of any single sub-task dominating the overall training process and leading to more robust model performance across all tasks.

\subsection{Cloud-edge collaboration}\label{appendexsec:cloud-edge}

\begin{figure}[htbp]
\centering
\includegraphics[width=9cm]{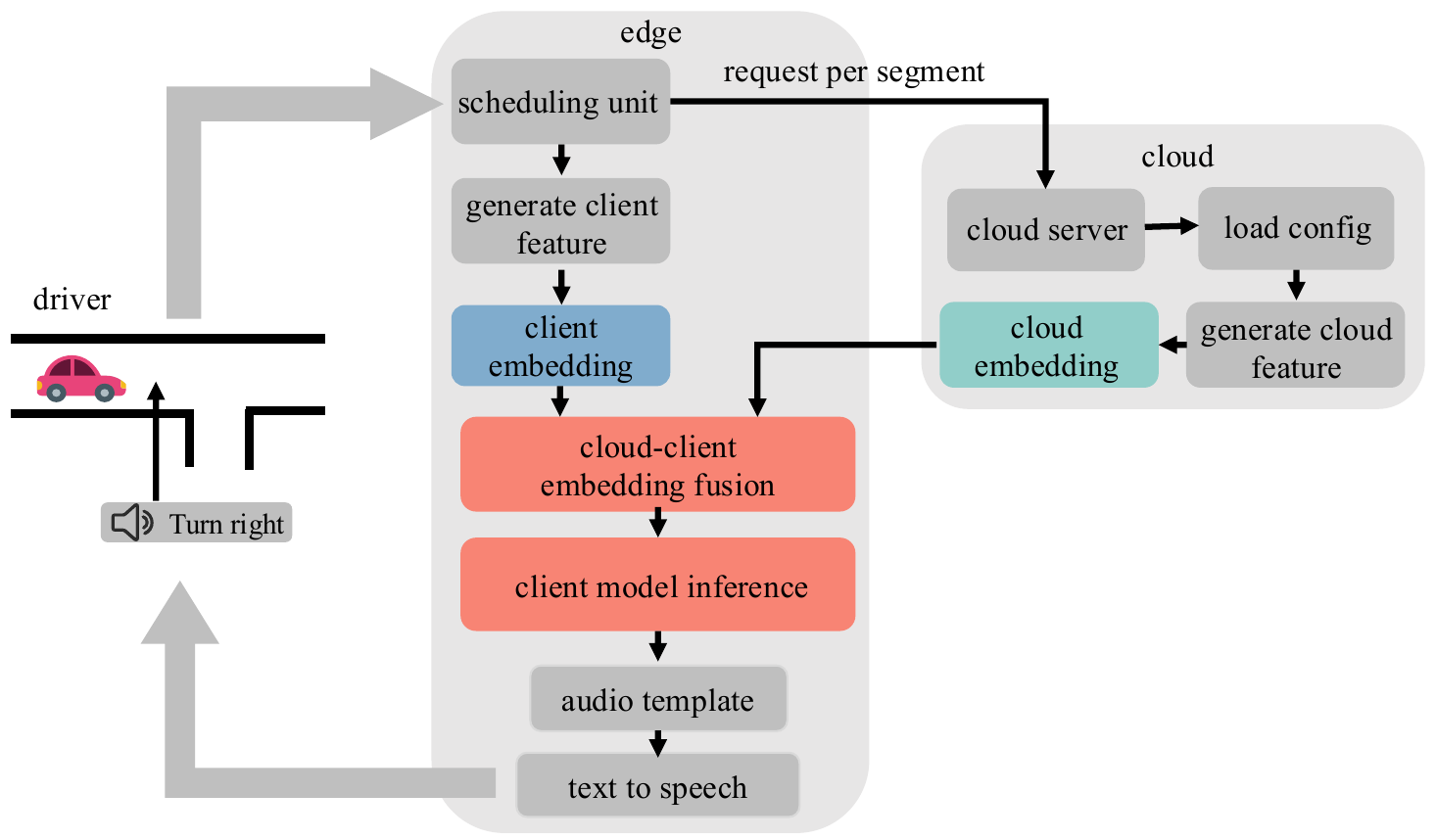}
\caption{\textbf{Cloud-edge collaboration framework.} The left side shows the user's driving behavior, and the edge device in the middle determines whether it needs to generate audio instruction based on the driving progress, as well as the start of each segment requesting cloud feature embedding from the cloud server on the right side.}
\label{fig:cloud_edge}
\end{figure}
As shown in Figure \ref{fig:cloud_edge}, we implement a cloud-edge collaborative architecture that cloud-edge collaboration framework for practical applications. This design leverages the strengths of both platforms to enhance real-time performance, system scalability, and driving experience.

By deploying the model inference on edge devices, we capitalize on the computational capabilities of user hardware. Processing data locally allows for real-time responsiveness, which is crucial for delivering timely audio instructions in navigation. It minimizes latency and ensures that drivers receive immediate feedback, enabling them to make quick and safe decisions on the road.

Offloading inference tasks to the edge also reduces the computational burden on the cloud servers. This not only decreases operational costs but also enhances the system's scalability by allowing it to support a larger user base without proportionally increasing cloud resources. It prevents the wastage of cloud server resources that would occur if all computations were centralized, especially considering that edge devices often have underutilized processing power.

While the edge devices handle real-time inference, the cloud server performs pre-processing tasks and generates embeddings for static and complex features such as road graph data, a part of element features, and personalized driver features. These computations benefit from the cloud's superior processing power and centralized data storage, which allows for up-to-date and comprehensive feature embeddings that can be periodically updated without impacting the edge devices. The scheduling unit on the edge is responsible for requesting the cloud-side model embedding for this segment from the cloud at the beginning of each segment and for orchestrating the edge-side play model inference.

An important advantage of this cloud-edge collaboration is the flexibility when updating the online model. With the inference model on the edge and feature embeddings on the cloud, we can update components independently. This modularity accelerates the iteration cycle of the model, reducing it from a monthly to a weekly timeframe. As a result, we can deploy updates and improvements more rapidly, responding promptly to user feedback and evolving requirements. This agility opens up greater possibilities for supporting additional features, enhancing the system's adaptability and longevity.

Moreover, differentiating the tasks based on their timing requirements optimizes system performance. Real-time processing is handled by the edge, meeting the immediate demands of navigation instructions. In contrast, the cloud handles tasks that can be pre-computed, like embedding updates, which do not require instant processing. This separation ensures efficiency by aligning computational tasks with the most suitable platform.

In conclusion, our cloud-edge collaborative approach effectively balances efficiency, effectiveness, and cost. By leveraging the computational strengths of edge devices for real-time inference and the cloud for intensive pre-processing tasks, we optimize resource utilization. The flexibility in updating the model enhances iterative efficiency, allowing for faster deployment of improvements and new features. This architecture not only improves the scalability and performance of the TBT navigation system but also significantly enhances the driver’s experience by providing timely, accurate, and context-aware audio instructions.

\section{Experiments}
In Section \ref{subsec:hyper}, we first introduce the dataset and model configurations. Subsequently, in Section \ref{subsec:ABtest}, we demonstrate the advantages of our approach through an AB test by deploying the model in real-world driving navigation and comparing it with our previously deployed HMM-based TBT audio instruction policy. Then, in Section \ref{subsec:ablation}, we evaluate the impact of key components of the model on the overall performance of the neural network through offline ablation experiments. Finally, in Section \ref{subsec:blindeval}, we randomly invited 100 drivers to participate in a blind evaluation of our model and the HMM-based TBT audio instruction policy. This evaluation covered 6 scenarios that are prone to yaw. The purpose of this assessment was to focus on the in-car experience of drivers in order to evaluate the effectiveness of our method on another dimension.

\subsection{Dataset and Model configurations}\label{subsec:hyper}
To train and evaluate our TBT audio instruction model, we construct a large-scale dataset derived from real-world driving navigation logs. We collect navigation trajectory data from actual drivers over 8 days, from June 11 to June 18, 2023. 

To uphold user privacy and maintain ethical integrity, all data was collected and processed in strict accordance with relevant data protection regulations established by the Chinese government and institutional ethical guidelines. Before data collection, informed consent was obtained from all participants, and comprehensive anonymization and aggregation procedures were implemented to eliminate any personally identifiable information. The resulting anonymized dataset was then employed to train and evaluate the proposed deep learning framework, ensuring that privacy safeguards were consistently upheld throughout the research process.

The navigation instruction policy for online data collection is based on a Hidden Markov Model (HMM) that we developed and deployed for TBT audio instruction. Unlike traditional distance-triggered methods that rely solely on geometric rules, our HMM-based policy models the sequential nature of driving states to improve instruction timing and content selection. This HMM policy serves as both the data collection mechanism and the baseline for comparison in our experiments. To maintain the quality and relevance of the collected data, we implement strict selection criteria during preprocessing. Specifically, we include only navigation paths initiated by cars and exclude trajectories with muted driver terminals, abnormal driving speeds, yawing, or GPS drifting during navigation. 

Additionally, we apply a secondary filter informed by domain knowledge to identify high-quality navigation trajectories. These trajectories are characterized by normal element transmission and appropriate audio playback timing within each segment. Beyond direct data collection, we utilize extra techniques to enhance the quality and diversity of the training dataset. Including filter out noisy or low-quality samples and based on real-road testing experiences, augment existing data by applying transformations such as voice trigger shifts or element perturbations to refine suboptimal samples into higher-quality ones. Furthermore, we create synthetic samples based on expert knowledge to address corner cases and represent challenging real-world scenarios.

These measures ensure that the dataset is both diverse and representative of complex driving conditions, allowing the model to generalize effectively across a wide range of scenarios. Ultimately, the filtered and enriched high-quality trajectories are used to construct the dataset for training the audio instruction model, ensuring its robustness and reliability in performance.

The datasets are partitioned into training, validation, and test sets. Feature standardization is performed using the mean and standard deviation calculated over the dataset. Sequential sample data are constructed by concatenating individual positioning point samples. The final dataset comprises approximately 1.56 billion sequence samples, with 1.1 billion samples in the training set (including 10 million supreme quality samples for supervised fine-tuning), 140 million samples in the validation set, and 320 million samples in the test set. This extensive dataset provides a robust foundation for training the model and assessing its performance in generating effective TBT audio instructions.

The model input features have 2139 dimensions and the length of sequence data is set to 3. The MoE part contains 3 experts. The model comprises 4 output heads: the trigger head outputs a scalar activated by the sigmoid function, the action head outputs a 28-dimensional vector also activated by the sigmoid function, the info head outputs a 25-dimensional vector activated by the softmax function, and the voice order head outputs a 5-dimensional vector activated by the softmax function. The model parameters are 1,147,943. During training, the learning rate is set to $0.001$, the batch size is $800$, and the model is trained for 400,000 steps, which takes approximately 38 hours on 8 NVIDIA T4 GPUs.

The optimization process for our model is conducted offline to ensure safety and reliability, as untested policies in real-world driving scenarios could result in suboptimal or dangerous navigation instructions. The offline training mode allows us to fully evaluate the model performance and risk to ensure the effectiveness and safety of navigation after deploy online.
Once validated, the model trained in float32 format is converted to float16 for deployment, which reduces memory usage while preserving performance. The model is subsequently divided into two components: a cloud component that manages static feature embeddings, and an edge component responsible for real-time inference on the user’s navigation device. After converting the cloud portion to ONNX, its size is optimized to approximately 807 KB for efficient inference on the cloud server. The edge component is further converted to MNN format \cite{alibaba2020mnn}, resulting in a model size of around 2.3 MB for deployment on the user's device. 
Once online, the model delivers TBT navigation services while continuously gathering new data from real-world usage. This data, which includes insights into corner cases and failure modes, is analyzed to refine the dataset and enhance the model’s robustness through subsequent offline retraining cycles. Because of the modularity of our cloud-edge architecture, this iterative process has been accelerated from monthly updates in HMM to a weekly schedule, facilitating rapid and continuous improvements.

\subsection{Real-world A/B Test}\label{subsec:ABtest}

To empirically validate the effectiveness of our model, we deploy the TBT audio instruction model online, in order to compare with the deployed HMM-based audio instruction policy. Our navigation system offers 4 modes to meet the different needs of drivers: detail, concise, minimalist, and intelligent. Drivers can select these modes based on their preferences. The detail and intelligent modes have more frequent audio prompts and are suitable for navigating unfamiliar roads, while the concise and minimalist modes have fewer prompts and are ideal for familiar routes.

\begin{figure*}[htbp]
\centering
\subfloat[]{
\begin{minipage}[b]{0.32\textwidth}
\centering
\centerline{\includegraphics[width=6cm]{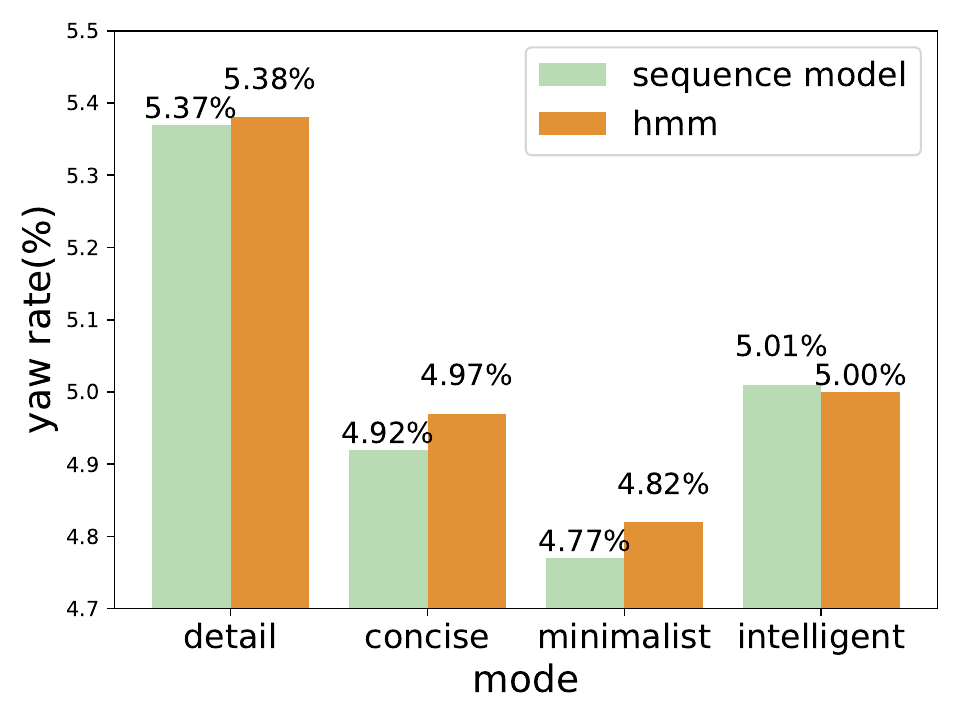}}
\label{fig:online::yaw_rate_online}
\end{minipage}
}
\subfloat[]{
\begin{minipage}[b]{0.32\textwidth}
\centering
\centerline{\includegraphics[width=6cm]{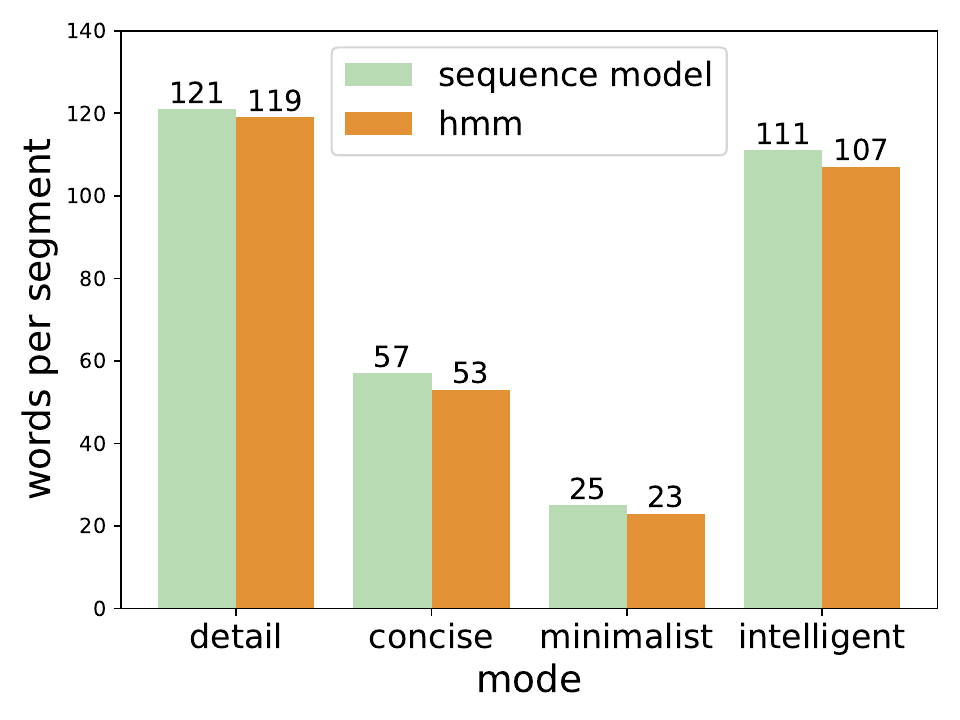}}
\label{fig:online::words_density_online}
\end{minipage}
}
\subfloat[]{
\begin{minipage}[b]{0.312\textwidth}
\centering
\centerline{\includegraphics[width=6cm]{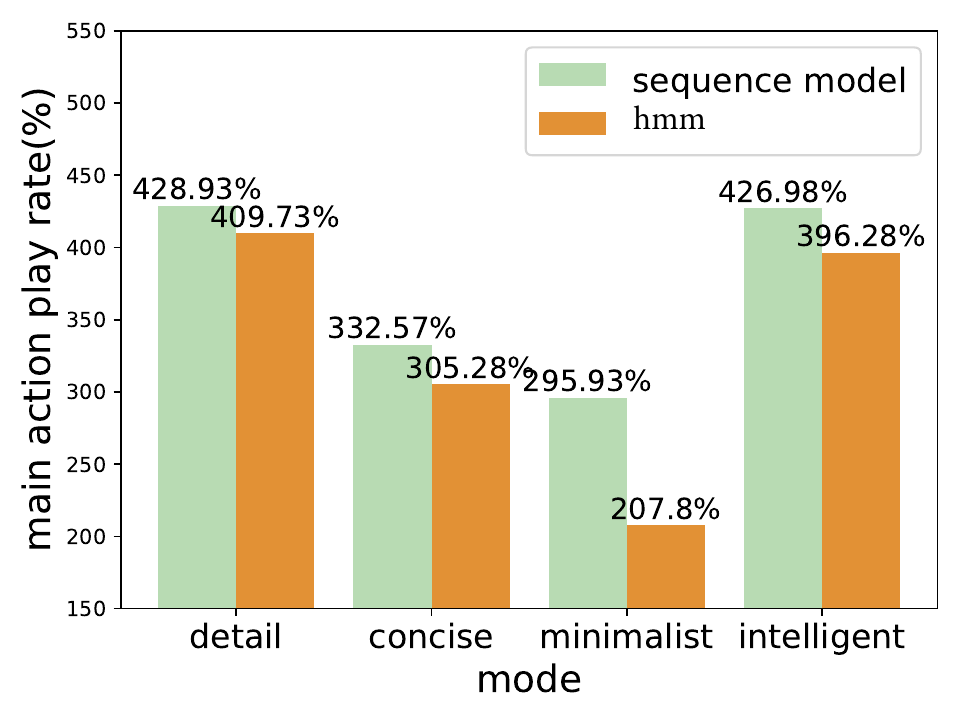}}
\label{fig:online::main_action_broadcast_rate_online}
\end{minipage}
}
\caption{\textbf{Real-world A/B test results.} Green represents the sequence model and orange represents the deployed HMM policy. (a) Yaw rate, lower yaw rates represent more effective audio instruction; (b) Words density, lower word density means more streamlined audio; (c) Main action element play rate, higher means more frequent TBT message alerts.}
\label{fig:online}
\vspace{-0.56cm}
\end{figure*}

The A/B test experiment period collected vehicle navigation data via our navigation system from August 28, 2024, to September 3, 2024, spanning a week and encompassing data from about 600 million segments. The primary results of the online experiment are illustrated in Figure \ref{fig:online}. The comparison mainly focuses on the yaw rate, the average words played per segment and the average play density of elements. The yaw rate is the ratio of yaw segments to all segments. Lower yaw rate means more accurate audio instruction. The play density of an element is define as the element play counts divided by the number of segments which can play this element. The average word count is positively related to the driver's difficulty in comprehending the content of the instruction audio, and the element play rate is positively related to the amount of information in the output content of the audio instruction model. In general, the more elements that are played with fewer average words represents a better content organization ability of the audio instruction model.

Compared to the deployed HMM-based baseline, our sequence model achieves a significant reduction in yaw rate. As shown in Figure \ref{fig:online} (a), except for the intelligent mode which has a slightly higher yaw, our model achieves a significant yaw rate reduction on all other modes, especially concise and minimalist. Note that since our daily online user volume is billions, $0.01 \%$ reduction in the yaw rate represents success in helping hundreds of thousands people drive to their destinations correctly. So the yaw improvement on the order of $0.01 \%$ is also significant.

Moreover, the increase in the average number of words played per segment, as illustrated in Figures \ref{fig:online}(b) and \ref{fig:online}(c), indicates that our model incorporates more main action elements with only 2-4 words increase. 
This indicates that our sequence model breaks through the seesaw effect of yaw rate, play density, and timing: with almost no increase in audio play words, the audio information density increase is achieved by significantly increasing the element play rates, and the impact on the play timing is few because the audio text length is almost unchanged. This in turn reduces the yaw rate.

In summary, the results of the real-world A/B test validate the effectiveness of our method in providing real-time, context-aware audio instructions that significantly reduce the yaw rate. In addition, our model can adapt to drivers' diverse navigation detail preferences, as evidenced by its superior performance in different modes, highlighting its robustness and generalizability under real-world driving conditions.

\subsection{Ablation Study}\label{subsec:ablation}
To investigate the necessity and effectiveness of each component in our model, we conducted a comprehensive ablation study offline by systematically removing or altering individual components and observing the impact on overall performance. The results are summarized in Table \ref{table:ablation}.
\begin{table}[H]
\vspace{-0.2cm}
\centering
\caption{Ablation study}
\vspace{-0.2cm}
\label{table:ablation}
\resizebox{0.5\textwidth}{!}{
\begin{tabular}{lccccc}
\hline
\textbf{Model Type} & \textbf{Trigger 10m} & \textbf{Trigger 30m} & \textbf{Action} & \textbf{Info} & \textbf{VoiceOrder}\\
\hline
our model & $83.3\%$ & $96.3\%$ & $97.0\%$ & $98.6\%$ & $90.7\%$\\
BERT decoder & $81.3\%$ & $96.2\%$ & $96.9\%$ & $98.3\%$ & $90.6\%$\\
LSTM & $81.6\%$ & $95.7\%$ & $96.5\%$ & $98.4\%$ & $90.0\%$\\
w/o position embedding & $80.8\%$ & $94.2\%$ & $96.8\%$ & $98.5\%$ & $88.8\%$\\
w/o MoE & $82.8\%$ & $96.2\%$ & $96.6\%$ & $98.4\%$ & $90.2\%$\\
w/o CrossNet & $80.3\%$ & $92.5\%$ & $92.9\%$ & $96.6\%$ & $88.6\%$\\
w/o sequence & $77.8\%$ & $94.9\%$ & $96.7\%$ & $98.4\%$ & $89.2\%$\\
\hline
\end{tabular}
}
\vspace{-0.2cm}
\end{table}

The metrics used in this section are defined as follows: \textbf{Trigger 10m}: The accuracy of the trigger head predictions within a 10-meter range; \textbf{Trigger 30m}: The accuracy of the trigger head predictions within a 30-meter range; \textbf{Action}: The accuracy of the action head predictions; \textbf{Info}: The accuracy of the info head predictions; \textbf{VoiceOrder}: The accuracy of the voice order head predictions.

Our model achieved a trigger 10m accuracy of $83.3\%$, a trigger 30m accuracy of $96.3\%$, an action accuracy of $97.0\%$, an info accuracy of $98.6\%$, and a voice order accuracy of $90.7\%$. These results affirm the robustness and high performance of our proposed method.

When we replaced the Decoder only Transformer with a BERT like Transformer, we observed a slight decrease in performance across all metrics. Specifically, trigger 10m and trigger 30m accuracies dropped by $2.0\%$ and $0.8\%$, respectively, while action, info, and voice order accuracies decreased marginally. This indicates that the Decoder only Transformer is better suited for capturing the spatial and temporal dependencies of sequence data in audio instruction tasks compared to BERT.

Replacing the Transformer with an LSTM also led to performance degradation across all metrics. This decline underscores the LSTM’s limitations in capturing long-range spatiotemporal dependencies and high-order feature interactions inherent in TBT navigation. While LSTM process sequences incrementally, their memory bottleneck and lack of parallelizable attention mechanisms hinder real-time adaptation to complex road conditions. In contrast, our Transformer’s self-attention design explicitly models global context and dynamic element relationships, enabling more precise audio instruction timing and content selection.

Removing the position embedding resulted in a more pronounced decline in performance, particularly for trigger 10m, trigger 30m, and voice order accuracies. This indicates that sequence and element position information is critical to the model prediction of timing and order.

Eliminating the MoE component led to a moderate reduction in performance, with the most significant impact observed on the action and voice order accuracies. This suggests that the multi-task learning capabilities of the MoE framework play a crucial role in effectively handling the diverse and interrelated sub-tasks of the TBT audio instruction model.

The removal of the CrossNet resulted in the most substantial performance degradation across all metrics, with Trigger 10m and Action accuracies decreasing by $3.0\%$ and $4.1\%$, respectively. This highlights the critical role of high-order feature interactions in capturing the complex relationships between different input features.

Finally, when we omitted the sequential input features, which means reducing the sequence length from 3 to 1, we observed a significant drop in Trigger 10m accuracy by $5.5\%$ and a noticeable decline in other metrics. This demonstrates the necessity of incorporating sequence information for providing accurate and context-aware audio instructions. 

\begin{table}[h]
\vspace{-0.2cm}
\centering
\caption{Sequence length ablation study}
\vspace{-0.2cm}
\label{table:seq_len}
\resizebox{0.5\textwidth}{!}{
\begin{tabular}{lccccc}
\hline
\textbf{Sequence length} & \textbf{Trigger 10m} & \textbf{Trigger 30m} & \textbf{Action} & \textbf{Info} & \textbf{VoiceOrder}\\
\hline
3 (our model) & $83.3\%$ & $96.3\%$ & $97.0\%$ & $98.6\%$ & $90.7\%$\\
1 & $77.8\%$ & $94.9\%$ & $96.7\%$ & $98.4\%$ & $89.2\%$\\
2 & $78.3\%$ & $94.3\%$ & $96.9\%$ & $98.6\%$ & $90.4\%$\\
4 & $83.4\%$ & $96.3\%$ & $97.1\%$ & $98.5\%$ & $90.7\%$\\
5 & $83.3\%$ & $96.3\%$ & $97.1\%$ & $98.6\%$ & $90.7\%$\\
\hline
\end{tabular}
}
\vspace{-0.2cm}
\end{table}
We conducted an ablation study to determine the optimal sequence length for our model by varying it from 1 to 5 and observing the impact on performance metrics. As presented in Table~\ref{table:seq_len}, reducing the sequence length to 1 and 2 resulted in a significant decline in performance. This indicates that shorter sequences fail to capture sufficient temporal dependencies, adversely affecting the model's ability to predict audio instruction timing within a critical 10-meter range. The minimal decreases in other metrics further underscore the importance of sequence data in accurately modeling spatiotemporal patterns essential for effective navigation instructions.

Conversely, increasing the sequence length beyond 4 yielded diminishing returns. Extending the sequence length to 5 shows a few improvements, but they are not substantial. Considering the elevated model inference costs associated with longer sequence lengths, a sequence length of 3 strikes an optimal balance between capturing adequate historical context and maintaining computational efficiency. This choice allows the model to effectively leverage spatiotemporal dependencies inherent in driving scenarios, enhancing the accuracy and contextual relevance of real-time audio instructions without incurring unnecessary computational overhead.

In conclusion, the ablation study confirms that each component of our model contributes significantly to its overall performance. The superior results achieved by our full model validate the design choices made during development and underscore the effectiveness of leveraging advanced deep learning techniques for real-time, context-aware instructions.

\subsection{Blind Evaluation}\label{subsec:blindeval}
To further assess the effectiveness of our proposed sequence model in real-world driving scenarios, we conducted a blind evaluation comparing our model's TBT navigation instructions with those generated by the deployed HMM-based policy. The goal was to evaluate the in-car experience from the driver's perspective, focusing on how well the instructions aid in navigating challenging driving situations that are prone to yaw.

\begin{figure}[htbp]
    \centering
    \subfloat[]{
    \begin{minipage}[b]{0.24\textwidth}
    \centering
    \centerline{\includegraphics[width=4cm]{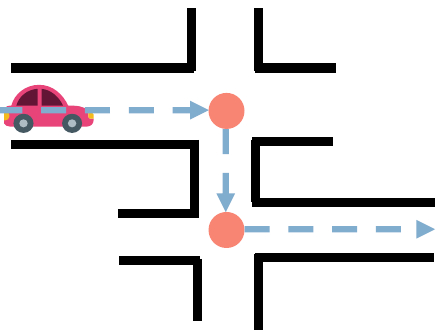}}
    \label{fig:scene::near}
    \end{minipage}
    }
    \subfloat[]{
    \begin{minipage}[b]{0.24\textwidth}
    \centering
    \centerline{\includegraphics[width=4cm]{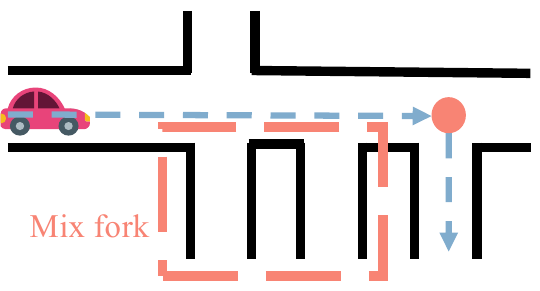}}
    \label{fig:scene::mixfork}
    \end{minipage}
    }\\\vspace{-0.4cm}
    \subfloat[]{
    \begin{minipage}[b]{0.24\textwidth}
    \centering
    \centerline{\includegraphics[width=4cm]{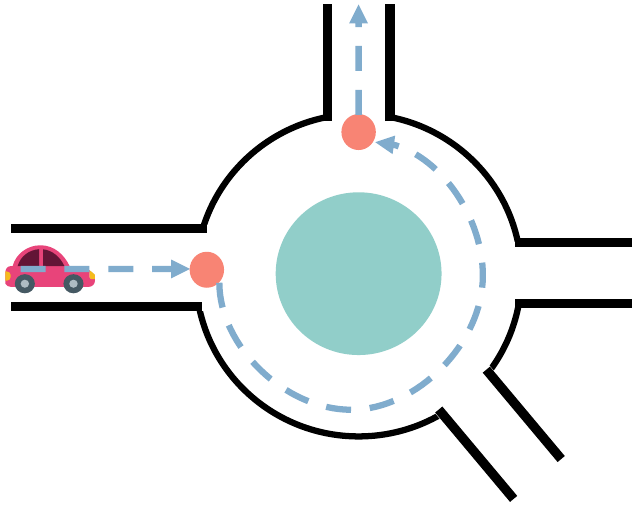}}
    \label{fig:scene::roundabout}
    \end{minipage}
    }
    \subfloat[]{
    \begin{minipage}[b]{0.24\textwidth}
    \centering
    \centerline{\includegraphics[width=4cm]{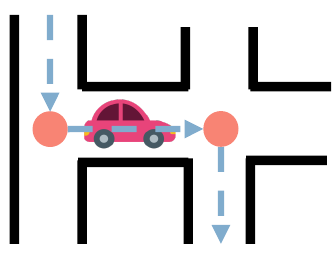}}
    \label{fig:scene::shortsegment}
    \end{minipage}
    }\\\vspace{-0.4cm}
    \subfloat[]{
    \begin{minipage}[b]{0.24\textwidth}
    \centering
    \centerline{\includegraphics[width=4cm]{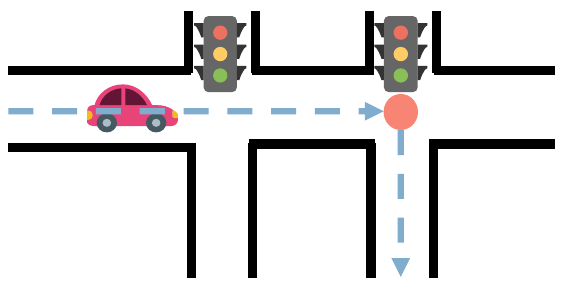}}
    \label{fig:scene::doublelight}
    \end{minipage}
    }
    \subfloat[]{
    \begin{minipage}[b]{0.24\textwidth}
    \centering
    \centerline{\includegraphics[width=4cm]{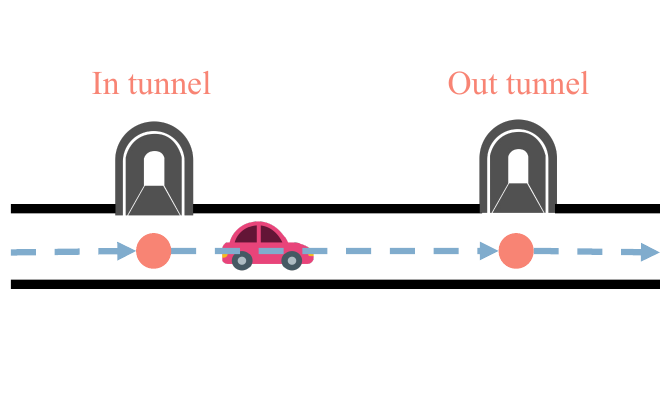}}
    \label{fig:scene::tunnel}
    \end{minipage}
    }
    \caption{\textbf{Blind eval scenarios.} (a) Near double bend; (b) Mix fork; (c) Roundabout; (d) Short segment; (e) Double traffic light; (f) Tunnel.}
    \label{fig:scene}
\end{figure}

We randomly recruited 100 drivers to participate in this study. Each driver was presented with pairs of instructions generated by our model and the HMM-based policy for six different driving scenarios known to cause navigational difficulties. We show these six scenarios in Figure \ref{fig:scene}:
\begin{itemize}
    \item \textbf{Near double bend.} Two consecutive turns ahead; pre-play next segment instructions to ensure smooth navigation and avoid incomplete voices due to short segment.

    \item \textbf{Mix fork.} Forks near the next navigation point may confuse drivers; warn against early turns when forks align with the upcoming turn direction.

    \item \textbf{Roundabout.} Circular roadway with a central island; drivers must follow consistent circulation and exit correctly, guided by audio instructions for the desired fork.

    \item \textbf{Short segment.} Brief road segments risk overshadowing or delaying audio cues; condense instructions to ensure timely delivery before entering.

    \item \textbf{Double traffic light.} Two traffic lights in quick succession; emphasize turning at the second light and remind drivers of the sequence as they approach.

    \item \textbf{Tunnel.} GPS signal may be lost inside tunnels; preload cloud-based data beforehand and provide post-tunnel instructions within the tunnel for lane changes.
\end{itemize}

\begin{table}[h]
\vspace{-0.2cm}
\centering
\caption{Results of the blind evaluation comparing user preference and average last instruction reaction time (RT) in segment.}
\vspace{-0.2cm}
\label{table:blind} 
\resizebox{0.5\textwidth}{!}{
\begin{tabular}{lccccccc}
\hline
\textbf{Scenario} & \multicolumn{3}{c}{\textbf{User Preference}} & \multicolumn{2}{c}{\textbf{Audio start RT (s)}} & \multicolumn{2}{c}{\textbf{Audio end RT (s)}} \\
\cline{2-4} \cline{5-6} \cline{7-8}
 & \textbf{Our Model} & \textbf{HMM} & \textbf{No Diff.} & \textbf{Our Model} & \textbf{HMM} & \textbf{Our Model} & \textbf{HMM} \\
 & \textbf{Better (\%)} & \textbf{Better (\%)} & \textbf{(\%)} & \textbf{(s)} & \textbf{(s)} & \textbf{(s)} & \textbf{(s)} \\
\hline
Near double bend & $21\%$ & $12\%$ & $67\%$ & $11.5$ & $12.5$ & $8.6$ & $9.5$ \\
Mix fork & $28\%$ & $22\%$ & $50\%$ & $8.9$ & $9.6$ & $7.5$ & $8.2$ \\
Roundabout & $54\%$ & $18\%$ & $28\%$ & $11.4$ & $12.2$ & $8.9$ & $9.8$ \\
Short segment & $45\%$ & $15\%$ & $40\%$ & $9.9$ & $10.8$ & $8.6$ & $9.7$ \\
Double traffic light & $21\%$ & $17\%$ & $62\%$ & $9.9$ & $10.7$ & $8.2$ & $8.9$ \\
Tunnel & $11\%$ & $18\%$ & $69\%$ & $11.9$ & $10.0$ & $9.7$ & $7.8$ \\
\hline
\end{tabular}
}
\vspace{-0.2cm}
\end{table}

For each scenario, the drivers were asked to listen to the audio instructions without knowing which model generated them and to rate which one they preferred or if they found them equally effective. 
The table \ref{table:blind} presented outlines user preferences for each of the six scenarios, along with the average reaction times of drivers at the beginning and conclusion of the last audio instruction in the segment, comparing our model to the HMM. This data indicates the duration in seconds that has elapsed since the driver heard the start (or end) of the most recent instruction before carrying out the corresponding action. A suitable reaction time ensures that the driver has an adequate opportunity to perform the required action without losing track of the navigation guidance provided \cite{dalton2013driving}.

The blind evaluation results include subjective user preferences and average confirmation reaction times for the final action element in each scenario. In challenging situations like the \emph{roundabout} and \emph{short segment}, drivers significantly preferred our sequence model's instructions over the HMM-based policy (
54\% vs. 18\% for roundabouts, and 45\% vs. 15\% for short segments). This preference is supported by shorter reaction times, which indicate that our model delivers the final action instruction closer to the upcoming turn within a navigation segment. This allows drivers to more easily visually identify the turn upon hearing the instruction, reducing cognitive load and enhancing situational awareness. This optimization is particularly significant in complex driving scenarios, such as roundabouts and short segments, further demonstrating the model's adaptability and practicality in dynamic environments.

In scenarios such as the \emph{near double bend}, \emph{mix fork}, and \emph{double traffic light}, most drivers perceived no significant difference, though a slight preference for our model emerged. The last audio start reaction times and end reaction times were marginally better for our model , indicating minor improvements in processing efficiency and user experience.

Conversely, in the \emph{tunnel} scenario, more drivers preferred the HMM policy (
18\% vs. 11\%). Our model's reaction time was longer than the HMM's, likely due to the signal-degraded environment. This suggests that our model's contextual richness may not always align with user expectations in sensor-compromised settings, pointing to an area for refinement.

Overall, the evaluation demonstrates that our model generally outperforms the HMM-based policy in delivering timely and contextually appropriate instructions, particularly in complex driving conditions. By balancing informational content with cognitive load and adapting to dynamic contexts, our model enhances driver situational awareness and decision-making. These findings validate the advantages of leveraging deep learning for TBT navigation systems.

\section{Conclusion}
In this paper, we introduce a novel deep-learning framework leveraging sequence models for real-time, context-aware audio instructions in TBT driving navigation. By formalizing the audio instruction generation into a multi-task learning problem and utilizing a cloud-edge collaborative architecture, our approach effectively balances informational content with cognitive load. Extensive experiments, including real-world A/B tests and blind evaluations, demonstrated that our method significantly reduces yaw rates compared to HMM-based policies, better conveying navigation instructions without overwhelming the driver. And the ablation studies confirmed the critical contributions of each component in our model.

Our method represents the first large-scale application of deep learning in practical TBT navigation systems, marking a substantial advancement in intelligent transportation technologies. Future work will focus on further optimizing model performance in complex scenarios and exploring personalized navigation experiences by integrating individual driver preferences and behaviors.

%
\bibliography{tbt_trans}
\bibliographystyle{IEEEtran}
\begin{IEEEbiography}[{\includegraphics[width=1in,height=1.25in,clip,keepaspectratio]{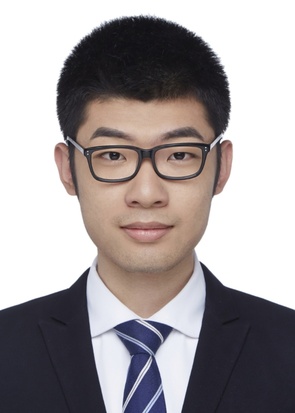}}]{Yiming Yang}
received the B.S. degree in electronic information engineering from the Beijing Institute of Technology, Beijing, China, in 2019, and the Ph.D. degree in control theory and control engineering from the Institute of Automation, Chinese Academy of Sciences, Beijing, China, in 2024.

He is currently an Algorithm Engineer with AMAP, Alibaba Group, Beijing, China. His research interests include deep learning for intelligent transportation systems, reinforcement learning, spatiotemporal sequence modeling, and agentic LLM.
\end{IEEEbiography}

\begin{IEEEbiography}[{\includegraphics[width=1in,height=1.25in,clip,keepaspectratio]{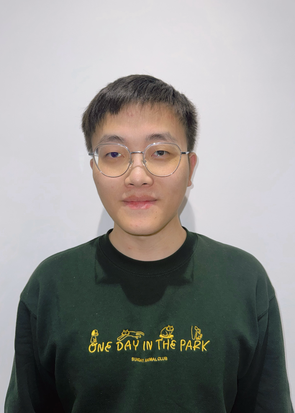}}]{Hao Fu}
received his master’s degree from the University of Science and Technology of China. He is currently with AMAP, Alibaba Group, Beijing, where he focuses on sequence prediction, intelligent transportation systems, and multi-agent systems. His research interests include artificial intelligence applications in transportation, spatiotemporal data modeling, and decision-making in dynamic environments.
\end{IEEEbiography}

\begin{IEEEbiography}[{\includegraphics[width=1in,height=1.25in,clip,keepaspectratio]{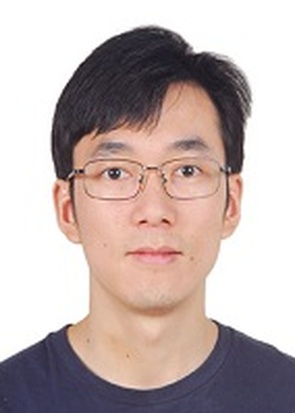}}]{Fanxiang Zeng}
received the B.S. degree in business management from the School of Economics and Management, Beijing University of Posts and Telecommunications (BUPT), Beijing, China, in 2010, and the Ph.D. degree in communication and information system from BUPT in 2017. From 2014 to 2016, he visited McGill University, Montreal, Canada, under the support of the China Scholarship Council as a visiting researcher under the supervision of Prof. M. D. Levine.

Currently, he is a Senior Algorithm Expert in AMAP, Alibaba Group. His research interests include visual tracking and detection, intelligent transportation systems such as  spatial-temporal modeling, turn-by-turn navigation, and agentic LLM and its applications in constrained travel planning and tour guiding areas.
\end{IEEEbiography}

\begin{IEEEbiography}[{\includegraphics[width=1in,height=1.25in,clip,keepaspectratio]{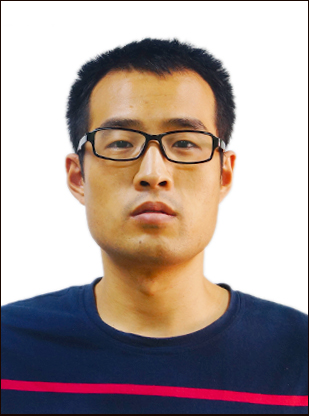}}]{Xikai Yang}
received his Master of Engineering degree in Computer System Architecture from Huazhong University of Science and Technology in 2013.
He joined AMAP in 2015, conducting research in geographic information science and spatial computing and led AMAP's spatial intelligence and intelligent navigation/positioning research in 2020.

He has been the director of AMAP's Terminal Intelligence Technology Department since 2024. His research interests include spatial intelligence, intelligent perception and interaction, intelligent navigation and positioning, intelligent terminals and edge computing, geographic information science and spatial computing.

\end{IEEEbiography}

\begin{IEEEbiography}[{\includegraphics[width=1in,height=1.25in,clip,keepaspectratio]{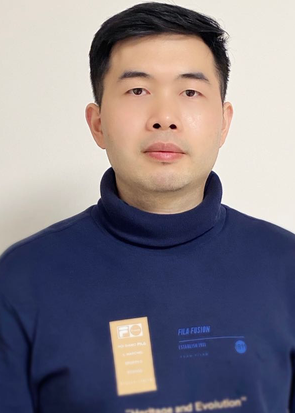}}]{Yue Liu}
earned both his bachelor's and master's degrees from Beijing Institute of Technology. He currently serves as a Director at AMAP, Alibaba Group where he leads the department of transportation and mobility intelligence. Prior to joining Alibaba, He worked at Search Technology Center Asia, Microsoft (STCA) until 2020, holding the position of Principal Engineering Manager.

His professional expertise spans spatio-temporal computing, natural language processing, dialogue systems, and speech technologies(ASR/TTS). Throughout his career, he has been dedicated to driving product and business innovation through cutting-edge technology. Some of his recent representative achievements include:
industry-first traffic signal inference based on spatio-temporal prediction (2022), Amap’s V2X warning system powered by spatio-temporal location computing (2023), Amap’s advanced environmental perception and pilot (2024), IP-based navigation voice (2024), and TrafficVLM (2025).
\end{IEEEbiography}

\begin{IEEEbiography}[{\includegraphics[width=1in,height=1.25in,clip,keepaspectratio]{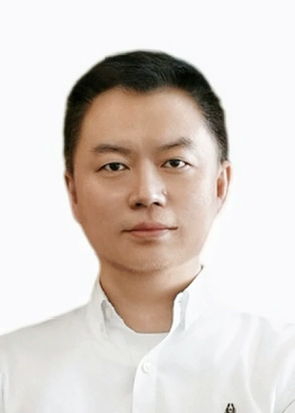}}]{Ning Guo}
received the master's degree from Zhejiang University in 2012. He has been the CEO of AMAP, Alibaba Group since 2024, defining AMAP as a spatial intelligence action engine built for the physical world.

Under his leadership, the team has independently developed AMAP’s world model, FantasyWorld, and pioneered several industry-first innovations, such as  
traffic signal inference based on spatio-temporal prediction, AMAP’s V2X warning system powered by spatio-temporal location computing, AMAP’s advanced environmental perception and pilot (AEP), the first local commerce recommendation system (AMAP Street Stars) based on human-place interaction modeling and computing, etc.
\end{IEEEbiography}

\vfill

\end{document}